\documentclass[11pt]{article}
\PassOptionsToPackage{numbers,square,sort&compress}{natbib}
\usepackage{acl}
\usepackage{times}
\usepackage{latexsym}
\usepackage[T1]{fontenc}
\usepackage[utf8]{inputenc}
\usepackage{microtype}
\usepackage{graphicx}
\usepackage{booktabs}
\usepackage{multirow}
\usepackage{amsmath}
\usepackage{url}

\title{PashtoCorp: A 1.25-Billion-Word Corpus, Evaluation Suite,\\
       and Reproducible Pipeline for Low-Resource Language Development}

\author{Hanif Rahman \\
  Independent Researcher \\
  \texttt{hanif@hanifrahman.com} \\
}

\begin{document}
\maketitle


\begin{abstract}
We present \textbf{PashtoCorp}, a 1.25-billion-word corpus for Pashto, a language spoken by
60 million people that remains severely underrepresented in NLP. The corpus is assembled
from 39 sources spanning seven HuggingFace datasets and 32 purpose-built web scrapers, processed
through a reproducible pipeline with Arabic-script tokenization, SHA-256 deduplication, and
quality filtering. At 1.25B words across 2.81 million documents, PashtoCorp is 40$\times$ larger
than the OSCAR Pashto subset and 83$\times$ larger than the previously largest dedicated Pashto
corpus. Continued MLM pretraining of XLM-R-base on PashtoCorp reduces held-out perplexity by
25.1\% (8.08$\to$6.06). On WikiANN Pashto NER, the pretrained model improves entity F1 by 10\%
relative (19.0\%$\to$21.0\%) and reduces training variance nearly 7$\times$; the largest gain appears
at 50 training sentences (+27\%), with PashtoCorp covering 97.9\% of WikiANN entity vocabulary.
On Belebele Pashto reading comprehension, Gemma-3n achieves 64.6\% accuracy, the first
published LLM baseline for Pashto on this benchmark. A leave-one-out source ablation shows
that Wikipedia (0.7\% of
documents) is the most critical source for NER: removing it alone reduces entity F1 by 47\%.
Corpus data, trained model, and code are available at
\url{https://huggingface.co/datasets/ihanif/pashto-corpus},
\url{https://huggingface.co/ihanif/xlmr-pashto}, and
\url{https://github.com/ihanif/pashto-corpus}.
\end{abstract}


\section{Introduction}

NLP has advanced rapidly over the past decade, but progress has been uneven across languages.
Pashto, an Indo-Iranian language spoken by 60 million people in Afghanistan, Pakistan, and
diaspora communities worldwide, remains poorly represented. The largest available Pashto
resources are the OSCAR 2301 subset ($\sim$31M words) and the NLPashto corpus ($\sim$15M words)
\cite{haq-etal-2023-nlpashto}. Both are too small for competitive language model training:
even the Pashto Wikipedia has only $\sim$6M words.

This paper makes three contributions:

\begin{enumerate}
\item \textbf{PashtoCorp}: a 1.25B-word corpus from 39 heterogeneous sources, built through a
fully reproducible pipeline.

\item An evaluation suite: MLM perplexity, vocabulary coverage, sample efficiency, and
baselines on POLD, WikiANN, and Belebele with fixed splits for reproducibility.

\item A source ablation: leave-one-out pretraining experiments quantifying each domain
group's contribution to language-model quality and downstream NER.

\end{enumerate}

Our results show that domain-adaptive pretraining improves performance when corpus and task
domains align (NER, news text) and does not when they diverge (POLD, social media). We report
null findings alongside positive results, as both provide actionable guidance for practitioners
working in low-resource settings.


\section{Related Work}

Prior Pashto NLP has been fragmented and small in scale. The NLPashto toolkit
\cite{haq-etal-2023-nlpashto} introduced a 15M-word corpus and a Pashto BERT model, with a
POS tagging model achieving 96.24\% accuracy on a 700K-word annotated corpus. The POLD
dataset \cite{ali-etal-2023-pold} provides 34,400 social media posts for offensive language
detection, with XLM-R (94.0\% F1) and Pashto BERT (94.3\% F1) baselines. WikiANN
\cite{pan-etal-2017-wikiann,rahimi-etal-2019-massively} includes a Pashto NER split with 300
sentences (100 each for train, validation, and test) and 7 BIO labels. Belebele
\cite{bandarkar-etal-2023-belebele} provides 900 Pashto reading comprehension questions with no
prior LLM baselines. The Universal Dependencies Pashto treebank \cite{nivre-etal-2016-universal}
has 155 annotated sentences.

Our HuggingFace sources include OSCAR \cite{ortiz-etal-2019-oscar}, CulturaX
\cite{nguyen-etal-2023-culturax}, CC-100 \cite{wenzek-etal-2020-ccnet}, MADLAD-400
\cite{kudugunta-etal-2023-madlad}, GlotCC \cite{kargaran-etal-2024-glotcc}, FineWeb2
\cite{penedo-etal-2024-fineweb}, and HPLT \cite{degibert-etal-2024-hplt}. Domain-adaptive
pretraining has been shown to improve downstream performance for multiple languages
\cite{gururangan-etal-2020-dont,nguyen-nguyen-2020-phobert}; gains depend on domain overlap
between pretraining data and evaluation tasks.


\section{Data Collection}

PashtoCorp is assembled from 39 sources: 7 HuggingFace datasets and 32 websites scraped with
purpose-built Scrapy spiders. Table~\ref{tab:sources} summarises the taxonomy.

\begin{table}[t]
\centering
\small
\caption{PashtoCorp sources by category.}
\label{tab:sources}
\resizebox{\columnwidth}{!}{%
\begin{tabular}{lrrr}
\toprule
\textbf{Category} & \textbf{Sources} & \textbf{Docs} & \textbf{Words (M)} \\
\midrule
Web crawl (HuggingFace) & 4 & 1,447,599 & 773.1 \\
News radio (scraped)    & 7 & 1,000,213 & 222.6 \\
Afghan news (scraped)   & 12 &  158,644  &  77.9 \\
Aggregator / blog       & 10 &  164,808  &  74.5 \\
PDF books / reports     &  1 &   16,867  &  81.6 \\
Encyclopedia            &  1 &   19,523  &  13.6 \\
Parallel / translation  &  2 &    1,592  &   4.7 \\
Other                   &  2 &    1,667  &   1.7 \\
\midrule
\textbf{Total} & \textbf{39} & \textbf{2,810,913} & \textbf{1,249.8} \\
\bottomrule
\end{tabular}}
\end{table}

\subsection{HuggingFace Datasets}
We incorporate seven datasets with Pashto or \texttt{pbt\_Arab} subsets: FineWeb2, HPLT~v2.0,
CulturaX, CC-100, MADLAD-400 (clean and noisy), and GlotCC-V1. Each is streamed and filtered
through our pipeline without storing intermediate files.

\subsection{Web Scraping}
We built 32 Scrapy spiders targeting Pashto news outlets and radio services. Major sources
include Azadi Radio (720,871 documents), VOA Pashto (205,284), BBC Pashto (51,736), Pajhwok
Afghan News (33,366), and Deutsche Welle Pashto (33,027).

Each spider has three fields: a start URL, a link-following rule
(e.g., \texttt{/archive/}, pagination patterns), and a CSS content selector
(e.g., \texttt{article p::text}). A \texttt{url\_must\_contain} path filter restricts
crawling to the language-specific URL path (e.g., \texttt{/pa/} for Pashto on RFE/RL and DW),
removing the need for an external language-ID model at crawl time. Sites with hashed CSS
class names (e.g., Next.js builds) use tag-based selectors (\texttt{article p::text})
rather than class-based ones. For JavaScript-rendered pages, a Playwright middleware
replaces the default HTTP downloader. A new spider typically requires 10--20 lines of
configuration.

To adapt the pipeline to another Arabic-script language (e.g., Dari, Urdu, Sindhi), only
the Unicode range in the language identification filter changes; spider configurations,
deduplication, and evaluation scripts carry over unchanged. For other scripts, the
character-range filter can be replaced with any token-level language detector.


\section{Processing Pipeline}

\subsection{Language Identification}
Each document is scored by the fraction of tokens in Pashto Unicode ranges
(U+0600--U+06FF and extensions). Documents with fewer than 70\% Pashto-script tokens are
discarded. This removes code-switched and mislabelled text while retaining documents with
numerals and Latin proper nouns.

\subsection{Deduplication}
Documents are deduplicated via SHA-256 hashes of lowercased, whitespace-normalised content.
This removes 356,943 duplicates (11.1\% of documents passing language identification).

\subsection{Quality Filtering}
Documents with fewer than 10 whitespace-separated tokens are removed. Table~\ref{tab:pipeline}
shows rejection rates per stage.

\begin{table}[t]
\centering
\small
\caption{Pipeline rejection statistics.}
\label{tab:pipeline}
\resizebox{\columnwidth}{!}{%
\begin{tabular}{lrr}
\toprule
\textbf{Filter stage} & \textbf{Docs removed} & \textbf{\% of raw} \\
\midrule
Pashto ratio $<$ 0.70 & 1,084,231 & 18.6\% \\
SHA-256 duplicate     &   356,943 &  6.1\% \\
Min tokens $<$ 10     &    42,817 &  0.7\% \\
\midrule
\textbf{Total rejected} & \textbf{1,483,991} & \textbf{25.5\%} \\
\textbf{Retained}       & \textbf{2,810,913} & \textbf{74.5\%} \\
\bottomrule
\end{tabular}}
\end{table}


\section{Corpus Analysis}

\subsection{Scale and Comparison}

PashtoCorp contains 1,249,765,401 words across 2,810,913 documents with 6,322,778 unique
word types. Table~\ref{tab:comparison} places it in context.

\begin{table*}[t]
\centering
\small
\caption{Comparison with prior Pashto corpora.}
\label{tab:comparison}
\begin{tabular}{lrrl}
\toprule
\textbf{Corpus} & \textbf{Words} & \textbf{Docs} & \textbf{vs.\ PashtoCorp} \\
\midrule
NLPashto \cite{haq-etal-2023-nlpashto} & $\sim$15M  &   ---  & 83$\times$ smaller \\
OSCAR 2301                             & $\sim$31M  &   ---  & 40$\times$ smaller \\
CC-100 (ps)                            & $\sim$54M  &  257K  & 23$\times$ smaller \\
MADLAD-400 (ps)                        & $<$1M      &    4K  & $>$1000$\times$ smaller \\
GlotCC (pbt-Arab)                      &   ---      &   37K  & --- \\
HPLT v2.0 (pbt-Arab)                   &   ---      &  466K  & --- \\
FineWeb2 (pbt\_Arab)                   &   ---      &  484K  & --- \\
\midrule
\textbf{PashtoCorp (ours)} & \textbf{1,250M} & \textbf{2,811K} & \textbf{1$\times$} \\
\bottomrule
\end{tabular}
\end{table*}

\subsection{Domain Distribution}

\begin{figure}[t]
\centering
\includegraphics[width=0.95\columnwidth]{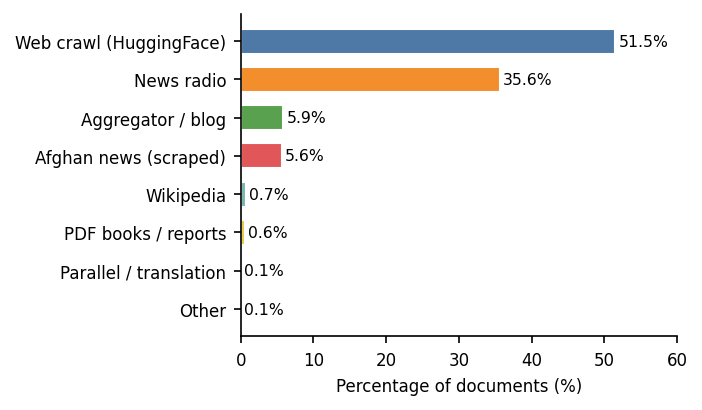}
\caption{Document distribution by domain category ($N=2{,}810{,}913$).}
\label{fig:domain}
\end{figure}

Figure~\ref{fig:domain} shows the domain distribution. Web crawl sources account for 51.5\%
of documents; news radio for 35.6\%. PDF books (0.6\% of documents, averaging 5,330 words each) are a register disproportionately
important for vocabulary (Section~\ref{sec:ablation}).

\subsection{Vocabulary Growth}

\begin{figure}[t]
\centering
\includegraphics[width=0.95\columnwidth]{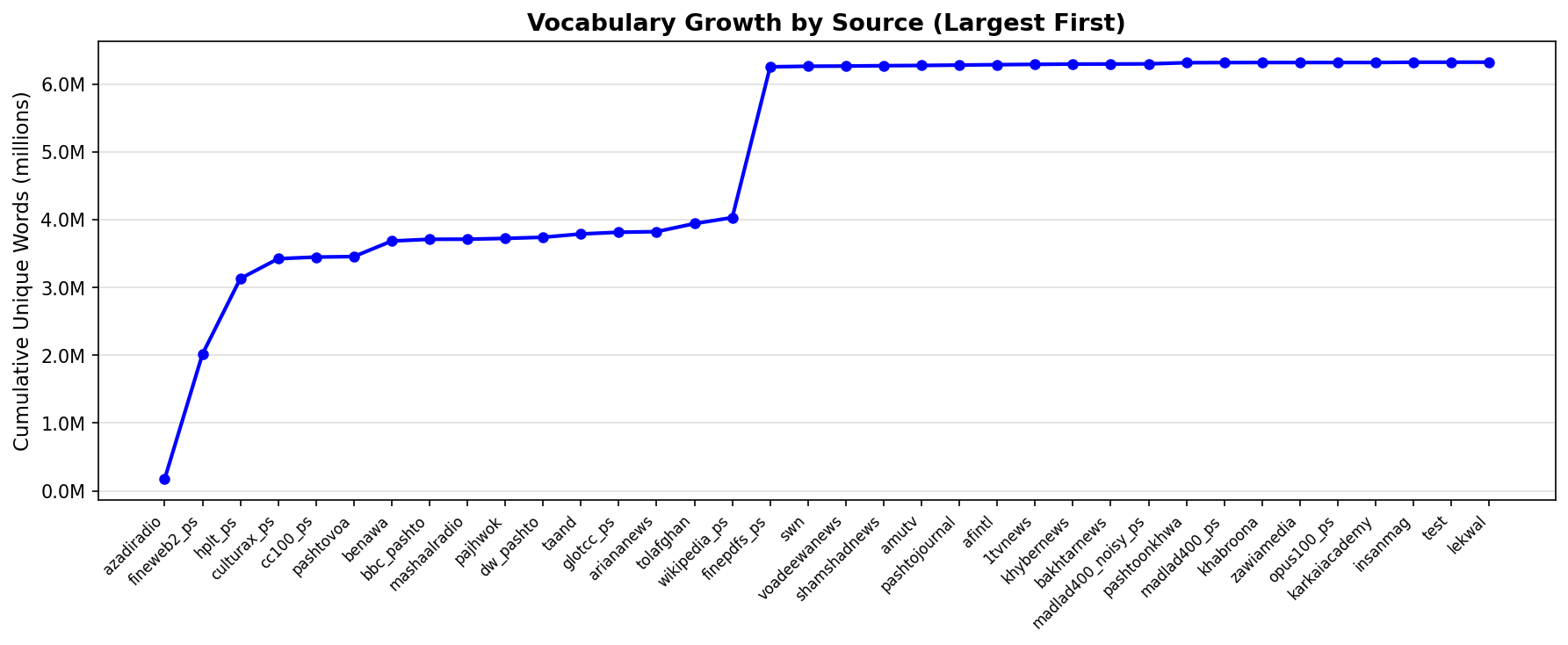}
\caption{Cumulative unique vocabulary (millions) as sources are incorporated in descending
document-count order. Despite its small size (16.9K documents, position 17), FinePDFs
contributes the largest marginal vocabulary of any single source: 2.22M unique types.}
\label{fig:vocab}
\end{figure}

Figure~\ref{fig:vocab} shows cumulative vocabulary as sources are added in descending document
count. FinePDFs (16,867 documents, 0.6\% of the corpus) adds 2,219,205 word types found in no
other source: 35\% of total vocabulary from less than 1\% of documents.

\subsection{Source Contribution Ablation}
\label{sec:ablation}

We conducted two leave-one-out ablations across six domain groups. The \emph{vocabulary
ablation} is a single-pass scan computing unique vocabulary and WikiANN entity coverage when
each group is excluded. The \emph{pretraining ablation} re-pretrains XLM-R-base on the
filtered corpus (100M-word cap, 400 gradient steps) for each group.

\paragraph{Vocabulary ablation.}
Table~\ref{tab:ablation_vocab} shows the results. FinePDFs contributes more unique vocabulary
than any other group despite covering only 0.6\% of documents: 2,219,205 word types from
16,867 digitized books and government reports. By contrast, News Radio accounts for 35.6\% of documents but adds only 55,234 unique types
(0.9\% of vocabulary); broadcast speech is lexically homogeneous. Entity coverage is robust across all ablations; even
removing the entire web crawl loses only 4 WikiANN entity tokens. Table~\ref{tab:ablation_vocab_source}
shows per-source marginal vocabulary contributions.

\begin{table*}[t]
\centering
\small
\caption{Vocabulary ablation (leave-one-out). Entity coverage = WikiANN tokens covered.}
\label{tab:ablation_vocab}
\begin{tabular}{lrrrr}
\toprule
\textbf{Group removed} & \textbf{Docs} & \textbf{Vocab} & \textbf{Vocab lost} & \textbf{Entity cov.} \\
\midrule
--- (Full corpus)          &         --- & 6,322,778 &         --- & 99.5\% \\
\textbf{PDF Books}         & \textbf{16,867} & \textbf{4,103,573} & \textbf{$-$35.1\%} & 99.5\% \\
Web Crawl                  & 1,447,599 & 4,418,444 & $-$30.1\% & 98.5\% \\
Afghan News                &   126,867 & 6,132,742 &  $-$3.0\% & 99.5\% \\
Wikipedia                  &    19,523 & 6,239,562 &  $-$1.3\% & 99.5\% \\
News Radio                 & 1,060,498 & 6,267,544 &  $-$0.9\% & 99.5\% \\
SWN                        &    12,546 & 6,314,810 &  $-$0.1\% & 99.5\% \\
\bottomrule
\end{tabular}
\end{table*}

\begin{table}[t]
\centering
\small
\caption{Per-source marginal vocabulary (top 10 sources).}
\label{tab:ablation_vocab_source}
\resizebox{\columnwidth}{!}{%
\begin{tabular}{lrrr}
\toprule
\textbf{Source} & \textbf{Docs} & \textbf{Marginal vocab} & \textbf{\% corpus} \\
\midrule
finepdfs\_ps   &  16,867 & \textbf{2,219,205} & \textbf{35.1\%} \\
hplt\_ps       & 397,273 &   495,782          &  7.8\% \\
fineweb2\_ps   & 483,890 &   345,935          &  5.5\% \\
culturax\_ps   & 335,609 &   207,160          &  3.3\% \\
benawa         &  58,319 &   201,093          &  3.2\% \\
tolafghan      &  20,525 &   106,514          &  1.7\% \\
wikipedia\_ps  &  19,523 &    83,216          &  1.3\% \\
taand          &  29,616 &    39,582          &  0.6\% \\
bbc\_pashto    &  51,736 &    22,505          &  0.4\% \\
glotcc\_ps     &  28,514 &    21,125          &  0.3\% \\
\bottomrule
\end{tabular}}
\end{table}

\paragraph{Pretraining ablation.}
Table~\ref{tab:ablation_pretrain} shows the results. Web crawl is most critical for
language-model quality: removing it raises perplexity from 6.06 to 6.58. Wikipedia is most
critical for NER: its removal drops entity F1 from roughly 20\% to 10.3\%, a 47\% relative
loss, despite Wikipedia being only 0.7\% of documents. News Radio removal slightly
\emph{improves} MLM perplexity (6.20), because broadcast speech is repetitive and its removal
yields a more diverse corpus per gradient step; however, news radio still matters for NER
(removing it drops F1 to 13.0\%).

\begin{table}[t]
\centering
\small
\caption{Pretraining ablation (400 steps each, sorted by PPL). Full 750-step model shown as
reference.}
\label{tab:ablation_pretrain}
\resizebox{\columnwidth}{!}{%
\begin{tabular}{lrrr}
\toprule
\textbf{Group removed} & \textbf{Docs} & \textbf{PPL} & \textbf{NER F1} \\
\midrule
\textit{Full (750 steps)} & \textit{2,810,913} & \textit{6.055} & \textit{19.6\%} \\
\midrule
Web Crawl    & 1,363,314 & \textbf{6.581} & 18.9\% \\
Wikipedia    & 2,791,390 &         6.412  & \textbf{10.3\%} \\
SWN          & 2,798,367 &         6.399  & 13.6\% \\
PDF Books    & 2,794,046 &         6.384  & 12.9\% \\
Afghan News  & 2,684,046 &         6.314  & 21.5\% \\
News Radio   & 1,750,415 &         6.201  & 13.0\% \\
\bottomrule
\end{tabular}}
\end{table}

\subsection{Zipf's Law}

Log-log regression of frequency rank vs.\ frequency gives $\alpha = 1.624$ ($R^2 = 0.985$),
higher than typical English values ($\alpha \approx 1.0$--$1.2$). This reflects Pashto's rich
inflectional morphology, which produces many low-frequency surface forms.

\begin{figure}[t]
\centering
\includegraphics[width=0.95\columnwidth]{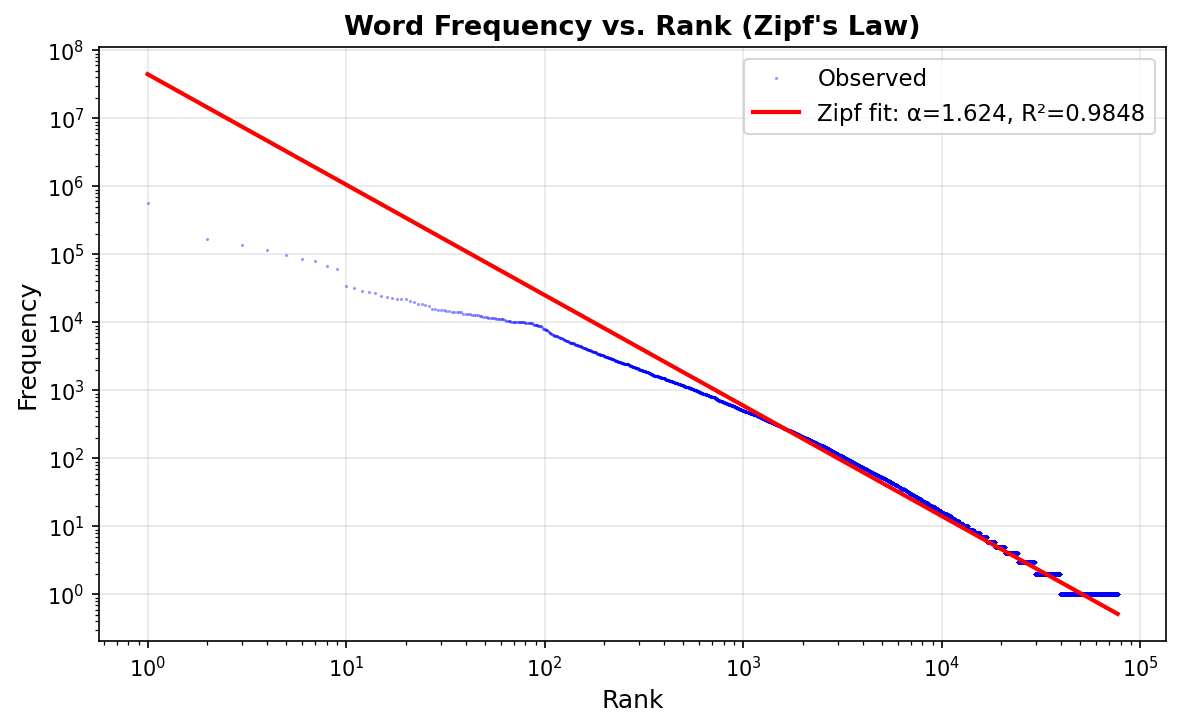}
\caption{Log-log frequency vs.\ rank for the top $10^5$ word types. The Zipf fit
($\alpha = 1.624$, $R^2 = 0.985$) holds closely.}
\label{fig:zipf}
\end{figure}


\section{Evaluation}

We evaluate PashtoCorp through intrinsic language-model quality metrics and extrinsic
downstream task performance via continued pretraining of XLM-R-base.

\subsection{Experimental Setup}

We continue MLM pretraining of \path{xlm-roberta-base} \cite{conneau-etal-2020-unsupervised}
on 100M words from PashtoCorp (283,569 training sequences of 512 tokens), using a 95/5
train/validation split. Training runs for 750 gradient steps with batch size 32, learning
rate $10^{-4}$, 200 warmup steps, and MLM probability 0.15, on Apple M4 (MPS/Metal) for
approximately 158 minutes. We refer to this model as \textbf{XLM-R+PashtoCorp}.

\subsection{Intrinsic: Perplexity and Vocabulary Coverage}

On 2,000 held-out PashtoCorp documents (3,204 sequences of 512 tokens), XLM-R-base scores
PPL~8.08 and XLM-R+PashtoCorp scores PPL~6.06, a 25.1\% reduction.

To understand the NER gains, we analyse overlap between PashtoCorp and WikiANN entity
vocabulary. PashtoCorp covers 97.9\% of entity tokens, higher than its 95.9\% overall coverage
(Table~\ref{tab:coverage}). Person names have 100\% coverage. Of entity tokens, 61.7\% span
multiple XLM-R subword units; in-domain pretraining directly improves representations for
these multi-token entities.

\begin{table}[t]
\centering
\small
\caption{PashtoCorp coverage of WikiANN tokens.}
\label{tab:coverage}
\resizebox{\columnwidth}{!}{%
\begin{tabular}{lccc}
\toprule
\textbf{Token set} & \textbf{Covered} & \textbf{Total} & \textbf{Coverage} \\
\midrule
All WikiANN tokens       & 2,062 & 2,151 & 95.9\% \\
Entity tokens (all)      &   381 &   389 & \textbf{97.9\%} \\
\quad PER (persons)      &   164 &   164 & 100.0\% \\
\quad LOC (locations)    &   101 &   102 &  99.0\% \\
\quad ORG (organisations)&   133 &   140 &  95.0\% \\
\bottomrule
\end{tabular}}
\end{table}

\subsection{Extrinsic: POLD Offensive Language Detection}

POLD \cite{ali-etal-2023-pold} contains 34,400 social media posts annotated for offensive
language. We apply an 80/10/10 stratified split (seed=42). XLM-R+PashtoCorp shows no
improvement over the baseline ($-$0.07pp F1, Table~\ref{tab:pold}). A vocabulary analysis
confirms the domain gap: PashtoCorp covers 91.6\% of POLD word types, 4.3pp below its
95.9\% coverage of WikiANN, with 3,651 OOV types that are colloquialisms and orthographic
variants specific to social media text.

\begin{table}[t]
\centering
\small
\caption{POLD results (macro-F1).}
\label{tab:pold}
\resizebox{\columnwidth}{!}{%
\begin{tabular}{lcc}
\toprule
\textbf{Model} & \textbf{Acc.} & \textbf{Macro-F1} \\
\midrule
CNN \cite{ali-etal-2023-pold}  & 92.4\% & 91.8\% \\
XLM-R-base (published)        & 94.8\% & 94.0\% \\
Pashto BERT (published)       & 94.8\% & 94.3\% \\
XLM-R-base (our repro)        & 94.6\% & 94.1\% \\
XLM-R+PashtoCorp              & 94.5\% & 94.0\% \\
\bottomrule
\end{tabular}}
\end{table}

\subsection{Extrinsic: WikiANN NER}

WikiANN Pashto \cite{pan-etal-2017-wikiann,rahimi-etal-2019-massively} provides 100/100/100
train/val/test sentences with 7 BIO labels. We fine-tune for token classification (10 epochs,
batch 16, lr=$5\times10^{-5}$, max 128 tokens) and evaluate with seqeval
\cite{nakayama-2018-seqeval} entity-level F1 over 5 seeds.

XLM-R+PashtoCorp reaches 21.0\% F1 (+10.3\% relative). Training variance drops from
$\pm$4.7\% to $\pm$0.7\%, a near 7$\times$ reduction (Table~\ref{tab:ner}).

\begin{table}[t]
\centering
\small
\caption{WikiANN Pashto NER (5 seeds, entity-level F1).}
\label{tab:ner}
\resizebox{\columnwidth}{!}{%
\begin{tabular}{lccc}
\toprule
\textbf{Model} & \textbf{F1 mean} & \textbf{F1 std} & \textbf{$\Delta$} \\
\midrule
XLM-R-base            & 19.0\% & $\pm$4.7\% & --- \\
XLM-R+PashtoCorp      & \textbf{21.0\%} & \textbf{$\pm$0.7\%} & +10.3\% \\
\bottomrule
\end{tabular}}
\end{table}

\paragraph{Per-entity-type breakdown.}
A separate run over 3 seeds measures F1 per entity type (Table~\ref{tab:ner_breakdown}).
PER benefits the most from pretraining (+11pp, +51\% relative), consistent with PashtoCorp
covering 100\% of WikiANN person name vocabulary. ORG and LOC see smaller improvements.
The base model's high variance on PER (15.6\%) reflects the instability of learning
person-name patterns from just 100 training sentences; pretraining more than halves this
standard deviation (5.8\%).

\begin{table}[t]
\centering
\small
\caption{NER per-entity-type F1 (3 seeds, mean $\pm$ std).}
\label{tab:ner_breakdown}
\resizebox{\columnwidth}{!}{%
\begin{tabular}{lcccc}
\toprule
\textbf{Model} & \textbf{Overall} & \textbf{PER} & \textbf{ORG} & \textbf{LOC} \\
\midrule
XLM-R-base       & 13.0\% $\pm$8.5\% & 21.8\% $\pm$15.6\% &  4.0\% $\pm$5.7\% &  9.6\% $\pm$9.1\% \\
XLM-R+PashtoCorp & \textbf{20.6\% $\pm$3.9\%} & \textbf{33.1\% $\pm$5.8\%} & \textbf{8.8\% $\pm$12.4\%} & \textbf{10.9\% $\pm$8.8\%} \\
\bottomrule
\end{tabular}}
\end{table}

\subsection{Sample Efficiency}

\begin{figure}[t]
\centering
\includegraphics[width=0.95\columnwidth]{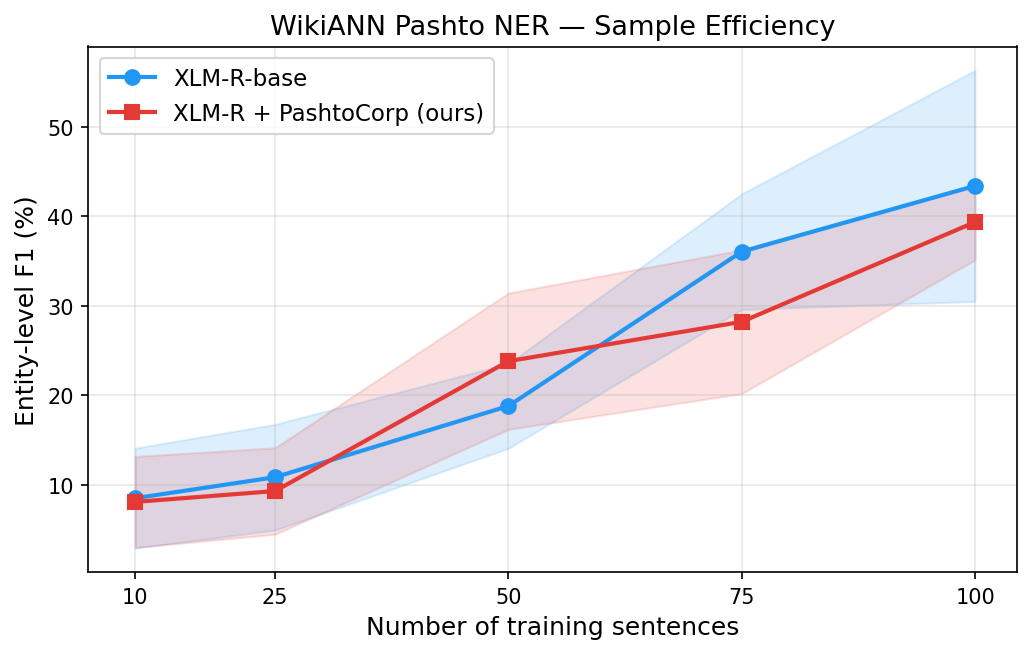}
\caption{WikiANN NER entity F1 vs.\ training set size (5 seeds, $\pm$1 std shaded).
PashtoCorp pretraining peaks at $n=50$ (+27\% relative F1).}
\label{fig:sample_eff}
\end{figure}

We vary training size $n \in \{10, 25, 50, 75, 100\}$ across 5 seeds per condition
(Table~\ref{tab:sample_eff}, Figure~\ref{fig:sample_eff}). The pretrained model shows the
largest relative advantage at $n=50$ (+27\% F1). At $n=100$ the base model's variance (12.9\%)
is 3$\times$ higher than the pretrained model's (4.2\%).

\begin{table*}[t]
\centering
\small
\caption{Sample efficiency: entity-level F1 by training size (5 seeds, mean $\pm$ std).}
\label{tab:sample_eff}
\begin{tabular}{lccc}
\toprule
\textbf{$n$} & \textbf{XLM-R-base} & \textbf{XLM-R+PashtoCorp} & \textbf{$\Delta$} \\
\midrule
10  & 8.5\%  $\pm$ 5.6\%  & 8.1\%  $\pm$ 5.1\%  & $-$0.4pp \\
25  & 10.9\% $\pm$ 5.9\%  & 9.3\%  $\pm$ 4.8\%  & $-$1.6pp \\
\textbf{50}  & \textbf{18.8\% $\pm$ 4.7\%} & \textbf{23.8\% $\pm$ 7.6\%} & \textbf{+5.0pp (+27\%)} \\
75  & 36.1\% $\pm$ 6.5\%  & 28.2\% $\pm$ 8.0\%  & $-$7.9pp \\
100 & 43.4\% $\pm$ 12.9\% & 39.4\% $\pm$ 4.2\%  & $-$4.0pp, $3\times$ lower $\sigma$ \\
\bottomrule
\end{tabular}
\end{table*}

\subsection{Belebele Reading Comprehension}

Table~\ref{tab:belebele} reports zero-shot 4-option multiple-choice accuracy on Belebele Pashto
(\texttt{pbt\_Arab}, 900 questions) \cite{bandarkar-etal-2023-belebele}. Encoder-based similarity methods perform at random chance; cosine similarity is not suitable
for multiple-choice comprehension. Accuracy rises consistently with model size within the Qwen3
family (27.7\%$\to$33.4\%$\to$37.9\% for 0.6B$\to$1.7B$\to$4B parameters), and
Llama-3.2-3B-Instruct reaches 40.3\%. Gemma-3n-E4B achieves 64.6\%, the first published LLM
result for Pashto on this benchmark; the remaining gap above 40\% likely stems from Gemma-3n's much wider multilingual
pretraining data.

\begin{table}[t]
\centering
\small
\caption{Belebele Pashto (zero-shot, 4-option MC).}
\label{tab:belebele}
\resizebox{\columnwidth}{!}{%
\begin{tabular}{lc}
\toprule
\textbf{Method} & \textbf{Accuracy} \\
\midrule
Random baseline                      & 25.0\% \\
MPNet embedding similarity           & 25.4\% \\
Pashto BERT embedding similarity     & 24.2\% \\
Qwen3-0.6B (Q8\_0, llama-cpp)        & 27.7\% \\
Qwen3-1.7B (Q8\_0, llama-cpp)        & 33.4\% \\
Qwen3-4B (Q4\_K\_M, llama-cpp)       & 37.9\% \\
Llama-3.2-3B-Instruct (Q4\_K\_M)     & 40.3\% \\
\textbf{Gemma-3n-E4B (UD-Q4\_K\_XL)} & \textbf{64.6\%} \\
\bottomrule
\end{tabular}}
\end{table}


\section{Discussion}

\paragraph{Domain alignment determines downstream gains.}
PashtoCorp pretraining benefits NER because corpus and task share overlapping text: news
articles and Wikipedia. It does not benefit POLD because conversational social media is absent
from the corpus. This is consistent with the domain-adaptive pretraining literature
\cite{gururangan-etal-2020-dont} and suggests that corpus builders should characterize domain
coverage before expecting improvements on a given downstream task.

\paragraph{Small, diverse sources have outsized value.}
The ablation results make a clear case for register diversity over raw volume. FinePDFs (0.6\%
of documents) contributes 35\% of total vocabulary. Wikipedia (0.7\% of documents) is the
single most important source for NER; its removal alone causes a 47\% relative F1 drop. For other low-resource corpus efforts, this is actionable: collecting encyclopedic and book
text early is disproportionately valuable, even at small scale.

\paragraph{Current pretraining uses a small fraction of PashtoCorp.}
Our experiments train on 100M words (400--750 gradient steps), approximately 8\% of the full
corpus. The 25.1\% perplexity reduction and 10\% NER improvement achieved in this limited
compute budget suggest substantially larger gains are possible with full-corpus training.

\paragraph{Reproducibility and transfer to other low-resource languages.}
Corpus assembly runs on a single machine in under 24 hours. Continued MLM pretraining
(750 steps, 100M words) took 158 minutes on an Apple M4 with no discrete GPU. The evaluation
suite (perplexity, vocabulary coverage, WikiANN NER, POLD, and Belebele) runs end-to-end from
publicly available checkpoints. Adapting the pipeline to another Arabic-script language
(e.g., Dari, Urdu, Sindhi) requires changing one Unicode range; adapting to any other script
requires one language detector swap. WikiANN and Belebele cover 282 and 122 languages
respectively, so the benchmarking harness transfers to any covered language without code
changes. Teams working on other under-resourced languages can apply the same data, model,
and evaluation stack with configuration changes only.


\section*{Limitations}

\paragraph{Pretraining scale.}
All pretraining experiments use 100M words (400--750 gradient steps), approximately 8\% of
PashtoCorp. The reported NER and perplexity gains should therefore be treated as lower
bounds on what the corpus can support with full training.

\paragraph{NER evaluation.}
WikiANN Pashto has only 100 training sentences; absolute F1 values are therefore
inherently unstable. The main NER result (19.0\%$\to$21.0\%) is reported over 5 seeds;
the source ablation NER values use a single seed and carry higher variance. We recommend
treating ablation NER values as relative indicators rather than precise estimates.

\paragraph{Corpus biases.}
PashtoCorp is 87.8\% news and web crawl text, predominantly from Afghanistan-based outlets.
Kandahari and Yousafzai Pashto are over-represented; Pakistani Pashto (Peshawar/Waziri
dialect) and diaspora register are under-represented. Models pretrained on PashtoCorp
may perform less well on text from these under-represented varieties.

\paragraph{POLD domain gap.}
PashtoCorp covers 91.6\% of unique POLD vocabulary types, compared to 95.9\% for WikiANN
(4.3pp lower). The 3,651 OOV POLD types include colloquialisms, informal verb forms, and
orthographic variants typical of social media that are absent from formal news text.
This coverage gap quantitatively explains the null pretraining result on POLD and limits
the scope of our extrinsic evaluation to news-adjacent tasks.


\section{Ethical Considerations}

All scraped content is publicly available text. No personally identifiable information beyond
what appears in published news is collected. PashtoCorp is predominantly news and web crawl
(87.8\% of documents), skewing toward Afghanistan-based sources and formal registers; dialect
coverage favours Kandahari and Yousafzai Pashto. PashtoCorp should not be used to build
systems that identify or surveil Pashto speakers, given the vulnerability of Pashto-speaking
communities in conflict-affected regions. POLD contains offensive content; downstream
moderation applications should include human oversight. Corpus data is released under the most
permissive license compatible with each source's terms; code is MIT; model checkpoints are
Apache 2.0.


\section{Conclusion}

We have presented PashtoCorp, a 1.25B-word Pashto text corpus assembled from 39 sources through
a reproducible pipeline. MLM pretraining on PashtoCorp reduces held-out perplexity by 25.1\%
and improves WikiANN NER by 10\% relative, with a near 7$\times$ reduction in training
variance. The pretrained model covers 97.9\% of WikiANN entity vocabulary, and the sample
efficiency analysis shows the largest gains at 50 training sentences. On Belebele, Gemma-3n
reaches 64.6\% accuracy, the first published LLM result for Pashto on this benchmark.

Two findings from the source ablation have broader relevance. Wikipedia and PDF books,
together fewer than 1.5\% of documents, are disproportionately valuable: removing Wikipedia
costs 47\% of NER performance; removing PDF books costs 35\% of vocabulary. For a new
low-resource language, collecting a small set of encyclopedic and book documents early is more
effective per document than scaling up web crawl data alone.

All data, code, checkpoints, and evaluation scripts are released: corpus at
\url{https://huggingface.co/datasets/ihanif/pashto-corpus}, model at
\url{https://huggingface.co/ihanif/xlmr-pashto}, and code at
\url{https://github.com/ihanif/pashto-corpus}.
Future work includes training on the full 1.25B-word corpus, developing Pashto-specific
generative models, and extending coverage to Pakistani Pashto and Waziri dialect.


\bibliography{paper}

\end{document}